\documentclass[letterpaper, 10 pt, conference]{ieeeconf}
\IEEEoverridecommandlockouts
\usepackage[utf8]{inputenc}
\usepackage[T1]{fontenc}
\usepackage{amsmath,amssymb,amsfonts}
\usepackage{booktabs}
\usepackage{graphicx}
\usepackage{microtype}
\usepackage{cite}
\usepackage{bm}
\usepackage{multirow}
\usepackage{colortbl}
\usepackage{xcolor}
\usepackage{algorithm}
\usepackage{algpseudocode}
\usepackage{url}
\usepackage{makecell}
\usepackage{enumerate}

\usepackage{amsthm}

\theoremstyle{plain}
\newtheorem{proposition}{Proposition}
\theoremstyle{remark}
\newtheorem{remark}{Remark}

\newcommand{\RTTableTwoRows}{%
Hanging Mug               & 8.0 & 2.0 & 6.0 & 13.0 & 5.0 & 8.0
                          & 30.0 & 26.0 & 19.0 & 20.0 & 6.0 & 8.0 \\
Open Microwave            & 10.0 & 9.0 & 31.0 & 4.0 & 37.0 & 15.0
                          & 58.0 & 14.0 & 77.0 & 61.0 & 57.0 & 44.0 \\
Place Can in Basket       & 34.0 & 38.0 & 43.0 & 44.0 & 30.0 & 29.0
                          & 69.0 & 75.0 & 74.0 & 78.0 & 52.0 & 60.0 \\
Place Shoe                & 41.0 & 47.0 & 62.0 & 68.0 & 42.0 & 40.0
                          & 64.0 & 57.0 & 66.0 & 70.0 & 63.0 & 53.0 \\
Move Playing Card Away    & 57.0 & 74.0 & 61.0 & 68.0 & 42.0 & 46.0
                          & 96.0 & 76.0 & 88.0 & 83.0 & 73.0 & 72.0 \\
  \multicolumn{13}{l}{$\cdots$ \emph{(50 tasks total)}} \\
Shake Bottle Horizontally & 87.0 & 96.0 & 99.0 & 99.0 & 91.0 & 95.0
                          & 98.0 & 100.0 & 100.0 & 100.0 & 98.0 & 100.0 \\
Pick Dual Bottles         & 58.0 & 52.0 & 47.0 & 42.0 & 34.0 & 25.0
                          & 95.0 & 97.0 & 80.0 & 90.0 & 67.0 & 81.0 \\
Place Bread in Basket     & 40.0 & 53.0 & 38.0 & 46.0 & 26.0 & 38.0
                          & 86.0 & 82.0 & 87.0 & 77.0 & 61.0 & 51.0 \\
Put Bottles in Dustbin    & 34.0 & 44.0 & 41.0 & 52.0 & 26.0 & 31.0
                          & 75.0 & 85.0 & 70.0 & 73.0 & 52.0 & 40.0 \\
Stack Blocks Two          & 19.0 & 11.0 & 71.0 & 83.0 & 34.0 & 47.0
                          & 96.0 & 94.0 & 99.0 & 99.0 & 75.0 & 67.0 \\
\midrule
  Overall Mean (50 Tasks) & 37.0 & 39.2 & 43.2 & 45.7 & 33.9 & 35.8
                                 & 73.8 & 66.9 & 71.7 & 70.8 & 57.7 & 59.3 \\
  $\Delta_s = Y_{s,2} - Y_{s,10}$ & \multicolumn{2}{c}{$+2.2$} & \multicolumn{2}{c}{$+2.4$} & \multicolumn{2}{c}{$+1.8$}
                                 & \multicolumn{2}{c}{${-}6.9$} & \multicolumn{2}{c}{${-}0.9$} & \multicolumn{2}{c}{$+1.6$} \\
  Tasks with $\Delta_s < 0$ & \multicolumn{2}{c}{21/50} & \multicolumn{2}{c}{19/50} & \multicolumn{2}{c}{20/50}
                                 & \multicolumn{2}{c}{36/50} & \multicolumn{2}{c}{16/50} & \multicolumn{2}{c}{20/50} \\
}

\title{\LARGE \bf Coupling Execution Horizon and Sampling Depth in Flow-Matching VLA Policies}

\author{%
Jiabing Yang$^{1,2,5}$, Yuan Xu$^{1,2}$, Yixiang Chen$^{1,2}$, Qisen Ma$^{1,2}$, Peiyan Li$^{1,2}$, Kai Wang$^{1,2}$,\\
Chenhang Cui$^{3}$, Tao Yu$^{1,2}$, Zichen Wen$^{4}$, Wen Ye$^{1,2}$, Bowen Fang$^{1,2}$, Ziheng He$^{1,2}$,\\
Xiangnan Wu$^{1,2}$, Yingda Li$^{1,2}$, Yan Huang$^{1,2,5,\dag}$, Liang Wang$^{1,2}$%
\thanks{$^{1}$School of Artificial Intelligence, University of Chinese Academy of Sciences.}%
\thanks{$^{2}$New Laboratory of Pattern Recognition (NLPR), Institute of Automation, Chinese Academy of Sciences. {\tt\small yangjiabing2025@ia.ac.cn, yhuang@nlpr.ia.ac.cn}}%
\thanks{$^{3}$National University of Singapore.}%
\thanks{$^{4}$Shanghai Jiao Tong University.}%
\thanks{$^{5}$FiveAges.}%
\thanks{$^{\dag}$Corresponding author.}%
}

\renewcommand{\baselinestretch}{0.952}
\begin{document}
\maketitle

\begin{abstract}
Flow-matching Vision-Language-Action (VLA) models generate expressive continuous trajectories for robotic manipulation, yet balancing real-time responsiveness with execution stability remains a persistent challenge. Standard deployment workflows typically treat numerical solver depth and execution horizon as decoupled choices. The horizon amortizes inference latency, while solver depth is kept high to preserve integration fidelity. In this paper, we reveal a fundamental compute--execution coupling in flow-matching VLAs governed by a systematic sign reversal. While coarse numerical integration degrades precision under rapid replanning, it markedly improves task performance under extended open-loop commitment. Geometrically, numerical truncation contracts lateral action dispersion around an invariant trajectory centroid, serving as an implicit regularizer that suppresses compounding open-loop drift. Across LIBERO, RoboTwin~2.0, and real-world Franka manipulation on representative VLA backbones including $\pi_0$, $\pi_{0.5}$, and GR00T, exploiting this joint coupling consistently achieves superior task success while drastically lowering inference latency, entirely training-free and without architectural modifications. These findings establish numerical sampling depth as an operational variance regulator, shifting its role from a static precision budget to an active deployment knob in robot learning. Project page: \url{https://jiabingyang01.github.io/Horizon-Depth-VLA/}.
\end{abstract}

\section{Introduction}
\label{sec:intro}

Flow-matching Vision-Language-Action (VLA) policies~\cite{black2024pi0,intelligence2025pi05,nvidia2025gr00t,black2025rtc,luan2026snapflow} generate continuous action trajectories~\cite{zhao2023act,chi2023diffusion} by numerically integrating learned velocity fields from prior Gaussian noise~\cite{lipman2023flow}. Deployment typically fixes two operational parameters, namely the \emph{execution horizon} $s$, the open-loop action prefix executed before re-querying the policy, and the \emph{sampling depth} $N$, the numerical integration steps allocated per chunk (equaling the number of function evaluations under forward Euler). For example, standard benchmark protocols for $\pi_0$ on LIBERO conventionally default to $(s,N)=(5,10)$.

Prevailing deployment practices treat $s$ and $N$ as independent choices. Receding-horizon frameworks~\cite{black2025rtc,liu2024bid} sweep execution horizons to amortize perception latency while holding solver depth fixed. Conversely, policy acceleration methods~\cite{prasad2024consistency,wang2024onedp,luan2026snapflow} distill or truncate numerical ODE solvers primarily under standard, short horizons. However, these two parameters are intrinsically linked. Solver depth $N$ dictates the dispersion of the generated action chunk, while execution horizon $s$ determines how long that dispersion compounds open-loop before sensory feedback intervenes.

\begin{figure}[t]
\centering
\includegraphics[width=\linewidth]{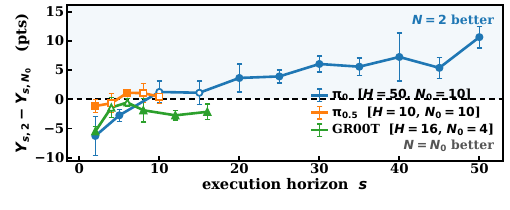}
\caption{Sign-reversing coupling between execution horizon $s$ and sampling depth $N$.
Shallow sampling ($N{=}2$) degrades performance under rapid replanning but yields systematic
gains under extended open-loop execution. Curves show paired marginal contrasts
$Y_{s,2} - Y_{s,N_0}$ on LIBERO-Long across matched seeds (mean $\pm$ sample SD), where
solid markers indicate horizons with unanimous sign agreement across all three seeds.}
\label{fig:phenom}
\end{figure}

In this work, we systematically investigate this interaction across frozen VLA checkpoints. On LIBERO-Long with $\pi_0$, the paired performance contrast between shallow ($N{=}2$) and deep ($N{=}10$) sampling exhibits a systematic sign reversal (Fig.~\ref{fig:phenom}). The reversal is consistent. While coarse solvers degrade task success under rapid replanning ($s{=}5$), they outperform deep solvers under extended open-loop execution ($s{=}20$). Multi-seed matched evaluations show that the contrast crosses zero between $s{=}5$ and $s{=}10$, shifting toward shallow sampling as the horizon lengthens. Experiments on RoboTwin~2.0 replicate this behavior on $\pi_{0.5}$ across bimanual tasks, while also highlighting boundary conditions governed by predictive horizons and baseline task capacity.

To understand this mechanism, we examine generated action distributions under matched initial noise. Forward Euler truncation contracts trajectory dispersion around an invariant centroid. Under rapid replanning, wider dispersion aids task progress as feedback quickly catches deviations; under extended open-loop execution, however, that dispersion compounds into drift. Here, the contracted distribution of a shallow solver acts as an effective implicit regularizer. The mechanism is geometric. Diagnostic interventions corroborate this account. Scalar noise rescaling fails to recover the effect and distorts mean trajectories, while freezing sampling noise across replans eliminates the deep solver's short-horizon advantage, which isolates redraw stochasticity as the primary driver.

\begin{figure*}[t]
\centering
\includegraphics[width=\textwidth]{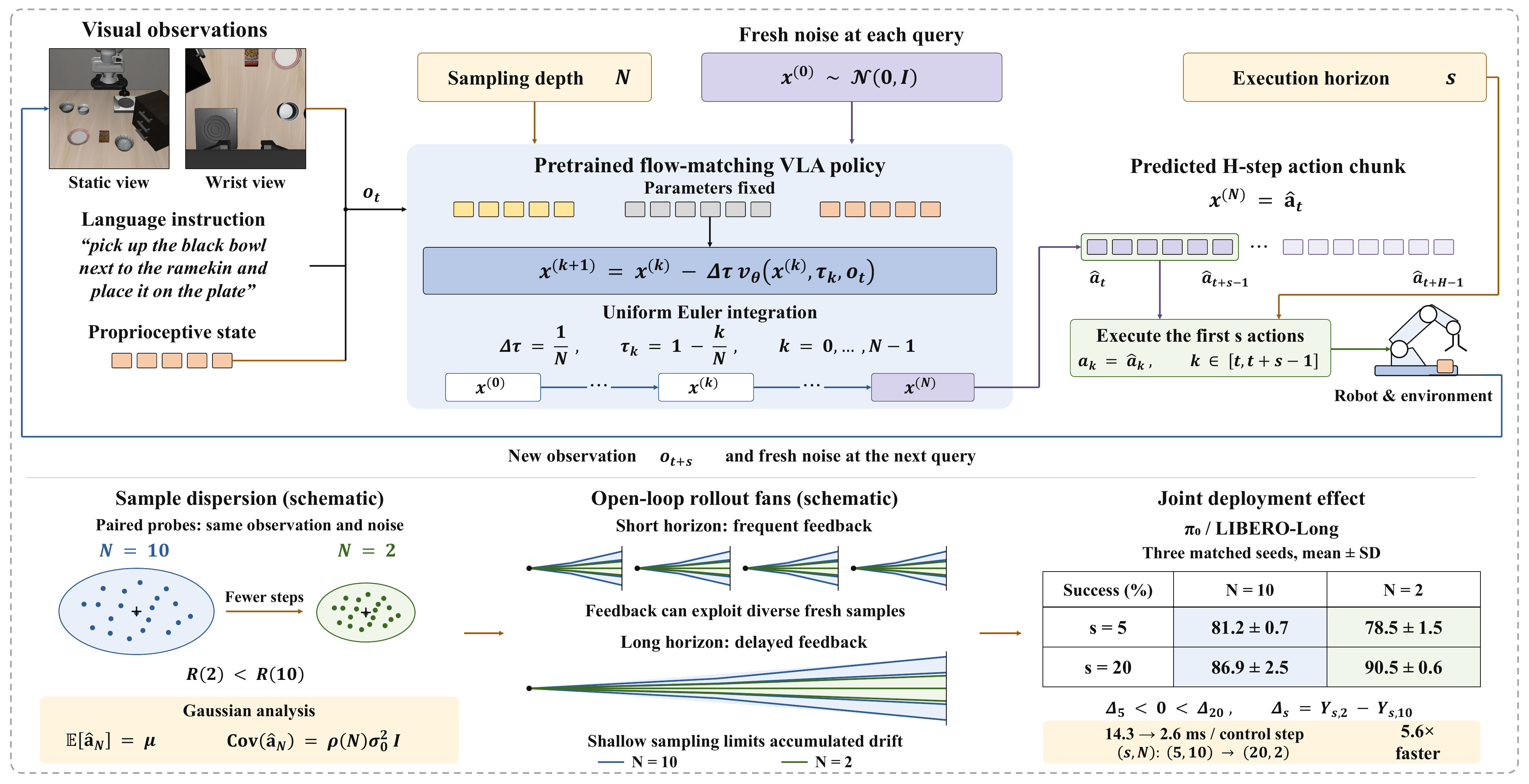}
\caption{Overview of computational-execution co-design in flow-matching Vision-Language-Action policies.
Top: The policy conditions on multimodal observation $o_t$ and integrates linear probability flow with $N$ forward Euler steps from prior noise $x^{(0)} \sim \mathcal{N}(0, I)$ to generate action chunk $\hat{\mathbf{a}}_t$, committing to an open-loop prefix of length $s$ on the robot before replanning at $t+s$.
Bottom: The coupling mechanism. Early solver truncation ($N{=}2$) contracts sample dispersion around the conditional centroid ($R(2) < R(10)$). Under rapid replanning ($s{=}5$), high-frequency closed-loop feedback exploits sample diversity; under extended commitment ($s{=}20$), coarse integration suppresses compounding lateral drift, reversing solver degradation and delivering up to $5.6\times$ inference speedups.}
\label{fig:pipeline}
\end{figure*}

This coupling provides practical and conceptual implications for robot learning. Practically, co-adjusting solver depth with execution horizon yields training-free Pareto improvements, improving task success while amortizing inference latency. Conceptually, it reframes numerical solver depth as an operational variance regulator that directly governs trajectory dispersion across the commitment window, offering a grounded perspective for compute--execution co-design (Figure~\ref{fig:pipeline}).

In summary, our key contributions are:
\begin{itemize}
    \setlength{\itemsep}{1pt}
    \setlength{\parskip}{0pt}
    \setlength{\topsep}{2pt}
    \item We identify a sign-reversing coupling between execution horizon and sampling depth in flow-matching VLAs: coarse solvers degrade performance under rapid replanning yet improve success under extended open-loop execution.
    \item We characterize the geometric mechanism: numerical Euler truncation contracts trajectory dispersion around an invariant centroid, serving as an implicit regularizer against compounding open-loop drift.
    \item We evaluate this phenomenon across simulation benchmarks (LIBERO, RoboTwin~2.0) and real-world Franka manipulation on representative VLA backbones, identifying operational boundaries for compute--execution co-design.
\end{itemize}

\section{Related Work}
\label{sec:related_work}

\noindent\textbf{Action Chunking and Horizon Scheduling.}
Chunked visuomotor policies trade feedback frequency for temporal smoothness~\cite{zhao2023act,chi2023diffusion}, underpinning representative flow-matching VLAs such as $\pi_0$~\cite{black2024pi0}, $\pi_{0.5}$~\cite{intelligence2025pi05}, and GR00T~\cite{nvidia2025gr00t}. To preserve reactivity, recent efforts introduce dynamic replanning mechanisms. Bidirectional decoding resamples candidate trajectories~\cite{liu2024bid}, reactive diffusion pairs slow planners with fast correction loops~\cite{xue2025reactive}, and VLA-Corrector triggers replanning from latent visual cues~\cite{pan2026corrector}. Other frameworks score trajectory prefixes via evaluation models~\cite{chen2026dynamic}, monitor chunk staleness with world models~\cite{chen2026dreamchunk}, train across variable horizons~\cite{jing2025mixture}, or overlap inference with asynchronous execution~\cite{black2025rtc}. Yet across these strategies, numerical sampling depth is held fixed at the backbone's default setting.

\noindent\textbf{Accelerated Sampling for Visuomotor Generation.}
Iterative numerical integration forms the primary latency bottleneck in diffusion and flow-matching policies~\cite{ho2020ddpm,song2021ddim,karras2022edm,lipman2023flow,liu2023rectified}. To accelerate inference, prevailing methods distill multi-step integration into few-step generators~\cite{prasad2024consistency,song2023consistency,wang2024onedp,luan2026snapflow}, stream partially denoised latents~\cite{hoeg2024sdp}, straighten velocity fields via MeanFlow~\cite{chen2026meanflow}, or adaptively truncate solver steps via error heuristics~\cite{hu2024adaflow}. While these methods primarily compress sampling budgets to lower inference latency, our work studies how sampling depth dynamically interacts with the execution horizon. We find that numerical solver truncation acts as an effective implicit regularizer under extended open-loop execution.

\noindent\textbf{Joint Computational and Execution Scheduling.}
Execution horizon and solver depth have traditionally been optimized in isolation. Reactive controllers tune replanning rates assuming fixed latency~\cite{black2025rtc,xue2025reactive}, while accelerated samplers compress per-query iterations under fixed commitments~\cite{prasad2024consistency,luan2026snapflow}. Recent work begins to couple compute and control dynamically: EQRL~\cite{wang2026eqrl} trains an RL meta-controller to adapt denoising steps, DVAC~\cite{feng2026dvac} monitors sampling variance to trigger early execution, and FASTER~\cite{lu2026faster} distributes compute non-uniformly across action tokens. While these frameworks design learned meta-controllers or variance monitors to allocate computational resources, our work demonstrates on frozen, unmodified backbones that numerical truncation naturally regularizes trajectory variance, offering a training-free foundation for compute--execution co-design.

\section{Methodology}
\label{sec:method}

\subsection{Flow-Matching Vision-Language-Action Policies}
\label{sec:flow_vla}

At control step $t$, the policy receives an observation $o_t$ comprising synchronized camera streams, proprioceptive state, and a language instruction, emitting an action chunk $\mathbf{a}_t = [a_t, \dots, a_{t+H-1}]^\top \in \mathbb{R}^{H \times D_a}$ spanning prediction horizon $H$ with control dimension $D_a$. Conditional flow matching synthesizes this chunk by transporting a tractable noise distribution onto the demonstration manifold~\cite{lipman2023flow,black2024pi0}. Under standard conventions~\cite{black2024pi0}, flow time $\tau{=}1$ corresponds to standard Gaussian source noise $p_1 = \mathcal{N}(0, I)$, $\tau{=}0$ corresponds to conditional target distribution $p_0(x \mid o_t)$, and the optimal transport displacement interpolant is:
\begin{equation}
x_\tau = (1 - \tau)x_0 + \tau x_1,
\qquad x_0 \sim p_0(x \mid o_t), \;\; x_1 \sim p_1 .
\label{eq:flow_path}
\end{equation}
A neural network $v_\theta(x_\tau, \tau, o_t)$ regresses the velocity field of this probability flow ODE, $\mathrm{d}x_\tau/\mathrm{d}\tau = v_\theta$.

Deployment integrates this probability flow backward from $\tau{=}1$ to $\tau{=}0$ starting from fresh sample $x^{(0)} \sim \mathcal{N}(0, I)$, discretized into $N$ uniform forward Euler steps of size $\Delta\tau = 1/N$ with time nodes $\tau_k = 1 - k/N$:
\begin{equation}
x^{(k+1)} = x^{(k)} - \Delta\tau\, v_\theta\!\left(x^{(k)}, \tau_k, o_t\right),
\label{eq:euler_update}
\end{equation}
where terminal state $x^{(N)}$ forms the emitted action chunk $\hat{\mathbf{a}}_t$. Each evaluation of $v_\theta$ requires a complete forward pass through the VLA backbone; thus $N$ strictly equals the number of function evaluations (NFE). While increasing $N$ reduces numerical discretization error, it scales per-query latency proportionally~\cite{song2021ddim,karras2022edm}, motivating standard practice to treat $N$ as an orthogonal computational budget.

\subsection{Receding-Horizon Action Chunking}
\label{sec:receding_horizon}

Executing an entire synthesized chunk open loop is fragile, as tracking errors compound without sensory feedback. Conversely, querying the heavy VLA backbone at every physical control step ($s=1$) incurs prohibitive latency and induces high-frequency motor chattering. Modern deployments therefore adopt receding-horizon execution~\cite{zhao2023act,black2025rtc}, combining periodic closed-loop replanning with piecewise open-loop commitment. When the policy predicts chunk $\hat{\mathbf{a}}_t = [\hat{a}_t, \dots, \hat{a}_{t+H-1}]^\top$, the controller commits to an open-loop prefix of length $s \in [1, H]$:
\begin{equation}
a_k = \hat{a}_k, \qquad k \in [t, t + s - 1],
\label{eq:open_loop_exec}
\end{equation}
triggering subsequent policy inference at step $t+s$ conditioned on fresh observation $o_{t+s}$. The execution horizon $s$ therefore defines the open-loop commitment window. Small $s$ provides frequent feedback to correct drift early, whereas large $s$ amortizes heavy backbone latency across physical steps at the cost of prolonged unguided rollout.

\subsection{Coupling Mechanism: Dispersion Contraction vs. Open-Loop Drift}
\label{sec:coupling_hypothesis}

As schematized in Figure~\ref{fig:pipeline}, standard deployment workflows typically decouple $N$ and $s$ as independent engineering choices. In physical reality, however, they are intrinsically coupled: $N$ governs how much prior stochastic dispersion survives into the generated chunk, while $s$ determines how long the robot commits open loop to that realization before closed-loop feedback intervenes.

During reverse probability flow integration from $\tau{=}1$ to $\tau{=}0$, the learned velocity field initially contracts prior noise toward the conditional mean trajectory ($\tau \approx 1$). Distinctions among plausible trajectory modes are resolved late in the flow ($\tau \to 0$), where probability density re-expands along feasible branches. A truncated coarse solver concentrates its evaluations in the initial contraction phase, capturing the conditional centroid while truncating late-stage lateral dispersion. The following proposition formalizes this contraction analytically on Gaussian flow paths.

\begin{proposition}[Euler Truncation Contraction]
\label{prop:euler_contraction}
Let $p_0 = \mathcal{N}(\mu, \sigma_0^2 I)$ and $p_1 = \mathcal{N}(0, I)$ be connected via linear flow path~\eqref{eq:flow_path}. The marginal velocity field is affine:
\begin{equation}
v(x, \tau) = \alpha(\tau)\bigl(x - (1-\tau)\mu\bigr) - \mu,
\label{eq:gauss_velocity}
\end{equation}
with gain $\alpha(\tau) = [\tau - (1-\tau)\sigma_0^2] / [\tau^2 + (1-\tau)^2\sigma_0^2]$.
For action chunk $\hat{\mathbf{a}}_N$ generated by $N$ forward Euler steps from $x^{(0)} \sim p_1$:
\begin{enumerate}[\normalfont (i)]
    \item \textbf{Centroid Invariance:} $\mathbb{E}[\hat{\mathbf{a}}_N] = \mu$ for all $N \ge 1$;
    \item \textbf{Covariance Contraction:} $\mathrm{Cov}(\hat{\mathbf{a}}_N) = \rho(N)\,\sigma_0^2 I$, where
    \begin{equation}
    \rho(N) = \sigma_0^{-2}\textstyle\prod_{k=0}^{N-1}\bigl(1 - \alpha(\tau_k)/N\bigr)^2 ,
    \label{eq:rho_def}
    \end{equation}
    satisfying boundary conditions $\rho(1) = 0$ and $\lim_{N \to \infty} \rho(N) = 1$;
    \item \textbf{Asymptotic Recovery Rate:} As $N \to \infty$, $\rho(N) = 1 - c(\sigma_0)/N + \mathcal{O}(N^{-2})$ with rate constant $c(\sigma_0) = 2 + \int_0^1 \alpha(\tau)^2\,\mathrm{d}\tau$.
\end{enumerate}
\end{proposition}

\begin{proof}
Along conditional mean trajectory $m(\tau) = (1-\tau)\mu$, $m(\tau) - (1-\tau)\mu \equiv 0$, nullifying the bracketed term in~\eqref{eq:gauss_velocity} so that $v(m(\tau), \tau) \equiv -\mu$. Because forward Euler integration is strictly exact for constant velocity fields, the mean integrates directly to $\mathbb{E}[\hat{\mathbf{a}}_N] = \mathbb{E}[x^{(0)}] - \sum_{k=0}^{N-1}\frac{1}{N}(-\mu) = \mu$ for all $N \ge 1$, establishing (i). Next, centered deviation $\delta^{(k)} = x^{(k)} - m(\tau_k)$ obeys linear recurrence $\delta^{(k+1)} = (1 - \alpha(\tau_k)/N)\delta^{(k)}$ with initial perturbation $\delta^{(0)} \sim \mathcal{N}(0, I)$, unrolling across $N$ steps to yield~\eqref{eq:rho_def}. At $N=1$, $\tau_0 = 1$ yields $\alpha(1) = 1$, which nullifies the factor $1 - \alpha(1) = 0$, causing boundary collapse $\rho(1) = 0$ and establishing (ii). Finally, noting $\alpha(\tau) = \frac{1}{2}\frac{\mathrm{d}}{\mathrm{d}\tau}\log[\tau^2 + (1-\tau)^2\sigma_0^2]$, its integral over $[0,1]$ evaluates to $-\log\sigma_0$. The leading term of $\log\rho(N) = 2\sum_{k} \log(1 - \alpha(\tau_k)/N) - 2\log\sigma_0$ cancels and $\rho(N) \to 1$. Expanding the Riemann sum via Euler-Maclaurin quadrature leaves endpoint overshoot $[\alpha(1) - \alpha(0)]/N = 2/N$ and second-order Taylor term $-\frac{1}{N}\int_0^1\alpha(\tau)^2\,\mathrm{d}\tau$, which exponentiates to the asymptotic rate in (iii).
\end{proof}

\begin{table*}[t]
\caption{Factorial Horizon--Depth Evaluation on the LIBERO Benchmark (Mean Success Rate \%).
For each model, we contrast the short-horizon pair ($s_{\mathrm{short}}$) and long-horizon pair ($s_{\mathrm{long}}$).
$\Delta_{\mathrm{short}}$ and $\Delta_{\mathrm{long}}$ denote the paired marginal effect of reducing solver depth;
Interaction $\mathcal{I} = \Delta_{\mathrm{long}} - \Delta_{\mathrm{short}}$ reflects the sign-reversing coupling across execution regimes ($\dagger$ denotes interaction contrasts constrained by baseline task ceiling effects).}
\label{tab:main_libero}
\centering
\setlength{\tabcolsep}{0.5mm}
\resizebox{\textwidth}{!}{%
\begin{tabular}{@{}l ccccc ccccc ccccc@{}}
\toprule
\multirow{2}{*}{Suite} & \multicolumn{5}{c}{$\pi_0$ \cite{black2024pi0} \scriptsize{[$H{=}50$]}}
& \multicolumn{5}{c}{$\pi_{0.5}$ \cite{intelligence2025pi05} \scriptsize{[$H{=}10$]}}
& \multicolumn{5}{c}{GR00T-N1.7 \cite{nvidia2026grootn17} \scriptsize{[$H{=}16$]}} \\
\cmidrule(lr){2-6}\cmidrule(lr){7-11}\cmidrule(l){12-16}
& $(5, 10)$ & $(5, 2)$ & $(20, 10)$ & $(20, 2)$ & $\mathcal{I}$
& $(5, 10)$ & $(5, 2)$ & $(10, 10)$ & $(10, 2)$ & $\mathcal{I}$
& $(2, 4)$ & $(2, 2)$ & $(8, 4)$ & $(8, 2)$ & $\mathcal{I}$ \\
\midrule
Spatial        & $97.1_{\pm 0.3}$ & $\mathbf{97.7}_{\pm 0.4}$ & $93.8_{\pm 0.5}$ & $95.3_{\pm 0.3}$ & $\mathbf{+0.9}_{\pm 0.6}$
               & $98.5_{\pm 0.4}$ & $98.5_{\pm 0.4}$ & $98.3_{\pm 0.6}$ & $\mathbf{98.7}_{\pm 0.6}$ & $\mathbf{+0.3}_{\pm 1.2}^{\dagger}$
               & $\mathbf{97.0}_{\pm 0.5}$ & $96.9_{\pm 0.6}$ & $96.7_{\pm 0.5}$ & $96.5_{\pm 0.4}$ & $\mathbf{-0.1}_{\pm 0.4}^{\dagger}$ \\
Object         & $98.1_{\pm 0.5}$ & $97.9_{\pm 0.8}$ & $97.1_{\pm 0.7}$ & $\mathbf{99.1}_{\pm 0.5}$ & $\mathbf{+2.1}_{\pm 1.5}$
               & $97.8_{\pm 0.3}$ & $98.5_{\pm 0.6}$ & $99.1_{\pm 0.6}$ & $\mathbf{99.5}_{\pm 0.2}$ & $\mathbf{-0.4}_{\pm 0.2}$
               & $97.1_{\pm 0.9}$ & $96.5_{\pm 0.8}$ & $98.9_{\pm 0.3}$ & $\mathbf{99.1}_{\pm 0.3}$ & $\mathbf{+0.7}_{\pm 0.4}$ \\
Goal           & $93.3_{\pm 1.0}$ & $93.9_{\pm 1.2}$ & $93.9_{\pm 0.1}$ & $\mathbf{97.3}_{\pm 0.6}$ & $\mathbf{+2.7}_{\pm 1.2}$
               & $98.3_{\pm 1.1}$ & $97.9_{\pm 0.5}$ & $98.0_{\pm 0.6}$ & $\mathbf{98.5}_{\pm 0.5}$ & $\mathbf{+0.8}_{\pm 0.5}$
               & $94.4_{\pm 0.5}$ & $94.4_{\pm 1.1}$ & $\mathbf{95.9}_{\pm 0.7}$ & $95.5_{\pm 0.5}$ & $\mathbf{-0.4}_{\pm 0.5}^{\dagger}$ \\
Long           & $81.2_{\pm 0.7}$ & $78.5_{\pm 1.5}$ & $86.9_{\pm 2.5}$ & $\mathbf{90.5}_{\pm 0.6}$ & $\mathbf{+6.4}_{\pm 2.1}$
               & $93.1_{\pm 0.4}$ & $93.1_{\pm 0.5}$ & $94.8_{\pm 1.0}$ & $\mathbf{95.3}_{\pm 0.8}$ & $\mathbf{+0.5}_{\pm 1.6}^{\dagger}$
               & $84.4_{\pm 0.3}$ & $78.9_{\pm 0.9}$ & $\mathbf{91.3}_{\pm 2.4}$ & $89.5_{\pm 0.8}$ & $\mathbf{+3.6}_{\pm 1.4}$ \\
\midrule
  Average & $92.4_{\pm 0.6}$ & $92.0_{\pm 0.5}$ & $92.9_{\pm 0.8}$ & $\mathbf{95.5}_{\pm 0.1}$ & $\mathbf{+3.0}_{\pm 0.8}$
                       & $96.9_{\pm 0.4}$ & $97.0_{\pm 0.2}$ & $97.6_{\pm 0.2}$ & $\mathbf{98.0}_{\pm 0.5}$ & $\mathbf{+0.3}_{\pm 0.5}^{\dagger}$
                       & $93.2_{\pm 0.3}$ & $91.7_{\pm 0.5}$ & $\mathbf{95.7}_{\pm 0.6}$ & $95.1_{\pm 0.3}$ & $\mathbf{+0.9}_{\pm 0.4}$ \\
\midrule
  Latency & $14.3$ ms & $10.2$ ms & $3.6$ ms & $\mathbf{2.6}$ ms & $\mathbf{5.6\times}$
                        & $14.2$ ms & $10.3$ ms & $7.1$ ms & $\mathbf{5.1}$ ms & $\mathbf{2.8\times}$
                        & $55.6$ ms & $39.6$ ms & $13.9$ ms & $\mathbf{9.9}$ ms & $\mathbf{5.6\times}$ \\
\bottomrule
\end{tabular}}
\end{table*}

\begin{remark}[Operational Variance Gain]
\label{rem:variance_gain}
While $\rho(1)=0$ reflects single-step collapse onto conditional expectation~\cite{liu2023rectified}, Proposition~\ref{prop:euler_contraction} formalizes the continuous rate $1 - c/N$ at which sampling depth governs dispersion. We define $g(N) \triangleq \sqrt{\rho(N)}$ as the \emph{operational variance gain}: injected noise scales by $g(N)$ in the emitted chunk. This framing yields clear empirical predictions, including centroid stability ($\delta_\mu \approx 0$ across all $N$), monotonic variance contraction scaling with $g(N)$ for $N \ge 2$, and structural divergence of single-step Euler ($N{=}1$), which breaks trajectory alignment and introduces field approximation error instead of structured variance contraction.
\end{remark}

Contraction in isolation neither helps nor hinders task completion; the execution horizon $s$ determines its operational value. Under frequent replanning ($s{=}5$), every query redraws the initial noise, and sensory feedback arrives before deviations compound. Here, diverse trajectory samples across redraws allow visual feedback to steer the robot around obstacles, while deep integration ($N{=}10$) supplies this dispersion. Under extended open-loop execution ($s{=}20$), the robot follows one draw for twenty steps without guidance. Here, wide dispersion compounds into physical drift, and coarse integration ($N{=}2$) stabilizes rollouts by removing excess variance while preserving the central path. This produces the sign reversal; reducing $N$ penalizes frequent replanning yet improves extended open-loop execution. The trade-off is direct.

\subsection{Geometric Diagnostics of Action Clouds}
\label{sec:diagnostic_framework}

We test Proposition~\ref{prop:euler_contraction} offline by assessing paired action cloud geometries on fixed visual observations. Passing an ensemble of identical Gaussian draws through shallow ($N{=}2$) and reference deep ($N{=}10$) solvers, we evaluate three complementary geometric metrics: \textit{Centroid Shift} $\delta_\mu = \|\bar{\mathbf{a}}_{N=2} - \bar{\mathbf{a}}_{N=10}\|_2$, with sample mean $\bar{\mathbf{a}}_N = \frac{1}{K}\sum_i \hat{\mathbf{a}}_i(N)$, measuring directional bias from truncation; \textit{Cloud Radius} $R(N) = \frac{1}{K} \sum_i \|\hat{\mathbf{a}}_i(N) - \bar{\mathbf{a}}_N\|_2$, with ratio $R(2)/R(10)$ measuring empirical variance contraction; and \textit{Linear Fit Residual}, measuring the fraction of shallow cloud variance unexplained by the optimal linear map of the deep cloud, diagnosing whether truncation alters geometry beyond scalar variance contraction.
\section{Experiments}
\label{sec:experiments}
We evaluate the horizon--depth coupling across 55 simulation tasks and five physical robot manipulation environments.

\subsection{Simulation Experiments}
\label{sec:exp_sim}

\noindent\textbf{Simulation Setup.}
To systematically evaluate policy dynamics across diverse manipulation settings, our benchmark encompasses three representative open-weight VLA models spanning distinct capacities: $\pi_0$ (3B)~\cite{black2024pi0}, a dual-system model coupling a PaliGemma-3B VLM backbone with a dedicated 300M action expert ($H{=}50, N_0{=}10$); $\pi_{0.5}$ (3B)~\cite{intelligence2025pi05}, an autoregressive visual-tactile token pipeline with a 10-action flow head ($H{=}10, N_0{=}10$); and GR00T-N1.7 (2B)~\cite{nvidia2026grootn17}, combining an Eagle-2B backbone with a diffusion action head ($H{=}16, N_0{=}4$). We evaluate these backbones across two standard benchmarks, namely the LIBERO benchmark~\cite{liu2023libero} spanning 40 tasks across four suites (Spatial, Object, Goal, and Long; 500 episodes per suite), and RoboTwin 2.0~\cite{chen2025robotwin} comprising 50 bimanual tasks evaluated in the official clean setting (100 matched rollouts per task). All rollouts run on NVIDIA A100 GPUs using official evaluation pipelines and released checkpoints.

\begin{figure}[t]
\centering
\includegraphics[width=0.96\linewidth]{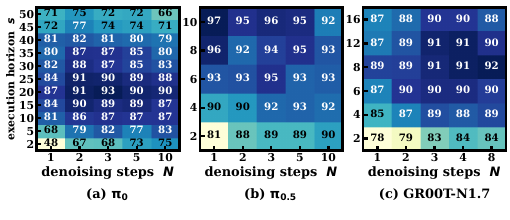}
\caption{Success rates across the joint execution horizon ($s$) vs. sampling depth ($N$)
grid on LIBERO-Long (500 rollouts per cell; $\pi_0$ and $\pi_{0.5}$ from seed 7, GR00T-N1.7 averaged over seeds 7/8/9; color scales normalized per panel).
(a) For $\pi_0$, optimal sampling depth shifts shallower as $s$ expands.
(b) $\pi_{0.5}$ is restricted by its 10-action prediction head.
(c) GR00T-N1.7 supervises 16 actions, peaking near its released configuration $(8,4)$.}
\label{fig:grid}
\end{figure}

\begin{table*}[t]
\caption{Horizon--Depth Evaluation on RoboTwin~2.0 across 50 Bimanual Tasks (Mean Success Rate \%)
evaluated across execution horizons ($s \in \{10, 25, 50\}$, $H{=}50$) and two sampling depths ($100$ rollouts per task).
Multi-task checkpoints are jointly fine-tuned across all 50 tasks.}
\label{tab:robotwin_summary}
\centering
\resizebox{\textwidth}{!}{%
\begin{tabular}{@{}l cccccc cccccc@{}}
\toprule
\multirow{2}{*}{Task Name} & \multicolumn{6}{c}{$\pi_0$ \scriptsize{[$H{=}50$]}}
& \multicolumn{6}{c}{$\pi_{0.5}$ \scriptsize{[$H{=}50$]}} \\
\cmidrule(lr){2-7}\cmidrule(l){8-13}
& $(10, 10)$ & $(10, 2)$ & $(25, 10)$ & $(25, 2)$ & $(50, 10)$ & $(50, 2)$
& $(10, 10)$ & $(10, 2)$ & $(25, 10)$ & $(25, 2)$ & $(50, 10)$ & $(50, 2)$ \\
\midrule
\RTTableTwoRows
\bottomrule
\end{tabular}}
\end{table*}

\noindent\textbf{LIBERO.}
Table~\ref{tab:main_libero} summarizes manipulation performance across all four suites of the LIBERO benchmark under our structured $2 \times 2$ factorial evaluation. Across heterogeneous environments and policy architectures, the empirical effect of numerical solver truncation fundamentally reverses across execution horizons. On LIBERO-Long with $\pi_0$, shallow sampling ($N{=}2$) slightly degrades success under the nominal reactive horizon ($s{=}5$), where high-frequency replanning exploits the directional diversity of deep integration. Once execution is extended to an open-loop commitment of $s{=}20$, however, the relationship systematically reverses; coarse sampling consistently outperforms the ten-step reference solver, raising success to $90.5\%$ with a robust positive interaction contrast of $+6.4\%$. Concurrently, amortizing inference over extended horizons (per-call median $45.93 + 2.57 N$\,ms on $\pi_0$, $70.5$/$51.4$\,ms on $\pi_{0.5}$, and $111.2$/$79.1$\,ms on GR00T-N1.7, divided by $s$) lowers per-step latency to $2.6$\,ms ($5.1$\,ms on $\pi_{0.5}$, $9.9$\,ms on GR00T-N1.7), unlocking up to $5.6\times$ deployment speedup alongside peak manipulation success. Tracking performance across the dense joint deployment grid (Figure~\ref{fig:grid}) and examining three-seed paired marginal differences $\Delta_s = Y_{s,2} - Y_{s,10}$ (Figure~\ref{fig:phenom}) localizes the crossover of the mean contrast between $s{=}5$ and $s{=}10$; the seeds disagree on its sign at $s{=}10$ and $15$, and it turns uniformly positive across all random seeds from $s{=}20$ onward for $\pi_0$. Architectural constraints naturally shape the accessible operational window. For $\pi_{0.5}$, the reachable execution horizon is strictly bounded by its 10-action prediction head ($s \le 10$), yet the marginal contrast exhibits the predicted reversal past $s{=}6$ before reaching chunk limits. Across all twelve evaluation blocks in Table~\ref{tab:main_libero}, the diagonal gain tracks distance below task ceiling with strong linear agreement ($r = 0.9$).

\noindent\textbf{RoboTwin 2.0.}
Table~\ref{tab:robotwin_summary} evaluates the factorial design across 50 bimanual manipulation tasks in the official clean setting, at three horizons of a 50-action chunk ($s \in \{10, 25, 50\}$). Bimanual manipulation substantially increases coordination complexity, as tasks divide between symmetric bilateral manipulation (e.g., \textit{Pick Dual Bottles}, \textit{Lift Heavy Tray}) requiring tightly synchronous arm velocities, and asymmetric handovers (e.g., \textit{Place Bread in Basket}, \textit{Hanging Mug}) where one arm stabilizes an articulated structure while the other performs fine insertion. On $\pi_{0.5}$, coarse sampling ($N{=}2$) incurs a notable performance penalty under rapid replanning ($s{=}10$), with 36 of 50 tasks losing ground. Extending the execution horizon to mid-chunk ($s{=}25$), however, completely eliminates this deficit. By full-chunk commitment ($s{=}50$), shallow sampling leads on 27 versus 20 tasks ($\Delta_{50} = +1.6\%$). This shift yields a statistically significant interaction contrast of $+8.5\%$ across tasks (95\% bootstrap CI $[+4.9\%, +12.0\%]$, $p < 10^{-4}$, positive on 37 of 50 tasks). This reversal is primarily driven by mitigating short-horizon degradation. Under frequent feedback, coarse sampling discards valuable exploratory diversity; under extended execution, suppressing lateral dispersion neutralizes compounding drift, preserving nominal trajectory execution across both arms. The dual-arm penalty vanishes.

\noindent\textbf{Pareto Frontier.}
Navigating along the co-design diagonal from the conventional reactive configuration $(10,10)$ to the balanced regime $(25,2)$ delivers a $3.5\times$ reduction in amortized step latency while maintaining robust manipulation success: $\pi_{0.5}$ preserves high overall performance ($70.8\%$ deduplicated, $70.7\%$ raw 8-shard sum, vs. $73.8\%$ at $(10,10)$), while $\pi_0$ improves from $37.0\%$ to $45.7\%$ ($45.6\%$ raw 8-shard sum, $+8.7$ points; $\mathcal{I}_{25} = +6.0\%$, 95\% CI $[+3.3\%, +8.8\%]$ on $\pi_{0.5}$, and $\mathcal{I} = -0.4\%$, 95\% CI $[-3.3\%, +2.4\%]$ on $\pi_0$). This establishes $s{=}25$ as an effective operational sweet spot that unlocks substantial compute savings while maintaining or improving task completion. For $\pi_0$, overall baseline success across these bimanual tasks is lower ($37.0\%$), and its short evaluation horizon ($s{=}10$) already sits past the policy's crossover point; the interaction contrast remains unresolved amidst baseline task difficulty. Under rapid closed-loop replanning, deep integration redraws a wide action cloud at every query, inducing oscillatory dual-arm chattering. In contrast, operating at $(25, 2)$ acts as a natural geometric low-pass filter over high-dimensional action space, stabilizing coordinated dual-arm rollouts.

\begin{figure}[t]
\centering
\includegraphics[width=\linewidth]{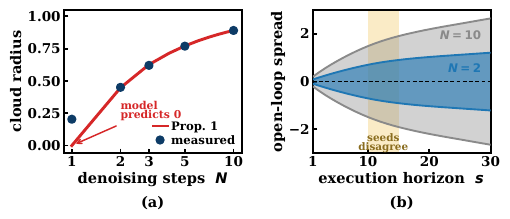}
\caption{Geometric mechanism of horizon--depth coupling.
(a) Cloud radius vs. sampling depth on $\pi_0$ (Proposition~1 curve fitted on $N \ge 2$).
Single-step Euler deviates by retaining a non-zero approximation residual.
(b) Open-loop trajectory spread accumulated over horizon $s$. Deep sampling ($N{=}10$)
generates $2.1\times$ wider lateral dispersion than $N{=}2$ while preserving nominal path direction.}
\label{fig:mech}
\end{figure}

\begin{table*}[t]
\caption{Causal Diagnostic Interventions on $\pi_0$/LIBERO-Long.
Success rates (\%) are mean $\pm$ sample SD across matched seeds $7/8/9$.
$\Delta_s = Y_{s,2} - Y_{s,10}$; interaction contrast $\mathcal{I} = \Delta_{20} - \Delta_5$.}
\label{tab:ablation_interventions}
\centering
\resizebox{\textwidth}{!}{%
\begin{tabular}{@{}l ccc ccc c@{}}
\toprule
\multirow{2}{*}{Experimental Condition} & \multicolumn{3}{c}{Short Horizon ($s{=}5$)} & \multicolumn{3}{c}{Long Horizon ($s{=}20$)} & \multirow{2}{*}{Interaction $\mathcal{I}$} \\
\cmidrule(lr){2-4}\cmidrule(lr){5-7}
& Deep ($N{=}10$) & Shallow ($N{=}2$) & Diff ($\Delta_5$)
& Deep ($N{=}10$) & Shallow ($N{=}2$) & Diff ($\Delta_{20}$) & \\
\midrule
  Standard Sampling (Default) & $81.2 \pm 0.7$ & $78.5 \pm 1.5$ & $-2.7 \pm 1.0$
                                     & $86.9 \pm 2.5$ & $90.5 \pm 0.6$ & $+3.7 \pm 2.4$
                                     & $\mathbf{+6.4 \pm 2.1}$ \\
\midrule
\multicolumn{8}{@{}l}{\textit{Intervention 1: 1D Dispersion Rescaling (Matched Endpoint Radius $\alpha_N$)}} \\
Scaled Noise Prior                   & $79.8 \pm 0.9$ & $70.9 \pm 1.7$ & $-8.9 \pm 2.5$
                                     & $88.3 \pm 1.3$ & $77.8 \pm 0.7$ & $-10.5 \pm 0.9$
                                     & $-1.5 \pm 1.6$ \\
\midrule
\multicolumn{8}{@{}l}{\textit{Intervention 2: Noise Freezing across Replans (Redraw Stochasticity Ablation)}} \\
Frozen Noise Source ($x^{(0)}_t \equiv x^{(0)}_0$) & $42.3 \pm 2.3$ & $72.9 \pm 0.6$ & $\mathbf{+30.6 \pm 2.5}$
                                                   & $75.0 \pm 1.7$ & $89.3 \pm 0.4$ & $\mathbf{+14.3 \pm 1.9}$
                                                   & $\mathbf{-16.3 \pm 4.3}$ \\
\bottomrule
\end{tabular}}
\end{table*}

\subsection{Mechanistic Analysis}
\label{sec:mechanistic}

\noindent\textbf{Geometric Contraction.}
To rigorously validate the analytical predictions of Proposition~\ref{prop:euler_contraction}, we examine the geometry of paired action clouds generated from identical visual inputs under frozen initial Gaussian noise vectors. Passing an ensemble of initial draws through varying integration depths confirms that the empirical cloud radius $R(N)$ closely mirrors the predicted asymptotic recovery curve $1 - c/N$ for all $N \ge 2$ (Figure~\ref{fig:mech}a); a single fitted mode width $\sigma_0 = 0.587$ reproduces every measured radius to within $0.006$. Across this entire range, the centroid shift never exceeds $\delta_\mu = 0.208$ (at $N{=}2$), $3.2\%$ of the nominal action magnitude, and the shallow centroid projects onto the deep-cloud mean at $1.000$. Early solver truncation thus preserves the directional orientation of the learned probability flow while selectively contracting lateral sample variance. Measuring the empirical cloud radius ratio yields $R(2)/R(10) = 0.51$; two integration steps halve the dispersion of the ten-step reference. Linear fit residual diagnostics, however, reveal that the best scalar map from each deep sample to its paired shallow sample leaves $48\%$ of the shallow-cloud variance unexplained, indicating that early truncation reshapes the covariance instead of scaling it isotropically; unimodality tests fail to reject a single mode on at least $85\%$ of observations at every depth, so the reshaping stays within one trajectory mode. Repeating this audit on a second $\pi_0$ bank (seed 9, nominal norm $7.50$) and two $\pi_{0.5}$ banks (seeds 8 and 9, nominal norm $3.92$) yields radius ratios of $0.51$ ($\delta_\mu = 3.1\%$) on $\pi_0$ and $0.43$--$0.45$ (per-observation mean ratios $0.34$ and $0.35$, $\delta_\mu = 2.3\%$, $53\%$ unexplained variance) on $\pi_{0.5}$, confirming that contraction holds across backbones.

Single-step Euler ($N{=}1$) fundamentally breaks this continuum. Instead of contracting variance along the learned probability flow, it incurs a substantial approximation residual that sits twice as far from the deep-cloud centroid as $N{=}2$ ($\delta_\mu = 0.405$) and retains no linear trace of its initial noise, with $99\%$ of its variance unexplained (scalar map slope $-0.02$); it introduces field approximation error instead of benign variance regularisation. This structural divergence explains why $N{=}1$ degrades performance across short and intermediate horizons: on LIBERO-Long across seeds $7/8/9$, single-step Euler falls $26.1 \pm 3.2$ points below the ten-step solver at $s{=}2$ ($48.7\% \pm 2.0\%$ vs. $74.8\% \pm 1.2\%$), trails two-step sampling at $s{=}20$ ($87.5\% \pm 0.4\%$ vs. $90.5\% \pm 0.6\%$), and only crosses zero past $s{=}35$ where coarse directional momentum aligns with the goal. At the trajectory level, unguided execution integrates single-query variance contraction over the commitment window $s$. Accumulating sampled action chunks into unguided rollouts reveals that the reference ten-step solver produces a $2.1\times$ wider lateral trajectory fan than $N{=}2$, close to the ratio $g(10)/g(2) = 2.0$ that Proposition~\ref{prop:euler_contraction} predicts from the single-query cloud radii alone, while center paths remain aligned within $3\%$ of total path length (Figure~\ref{fig:mech}b). Across extended horizons, the absolute lateral gap between deep and shallow trajectory fans widens. It reaches one unit of open-loop deviation near $s{=}16$, just before the seeds stop disagreeing on the sign of $\Delta_s$ at $s{=}20$.

\noindent\textbf{Causal Interventions.}
Table~\ref{tab:ablation_interventions} presents targeted causal interventions isolating the underlying driving mechanisms. To evaluate whether the empirical advantages of shallow sampling stem from generic variance reduction, we evaluate an endpoint-matched noise rescaling intervention that artificially scales initial noise variance. This intervention fails to reproduce the sign reversal. It degrades policy performance across both execution horizons, producing a negative interaction contrast ($\mathcal{I} = -1.5$). Rescaling the initial Gaussian prior forces trajectory integration through out-of-distribution velocity fields, nearly doubling centroid drift from $0.208$ to $0.382$. This confirms that the performance gains of $N{=}2$ originate from structured ODE contraction along the learned probability flow, which arbitrary variance manipulation cannot replicate.

Conversely, freezing the initial noise vector across successive replanning cycles isolates the role of stochastic redraw diversity. Under frequent closed-loop replanning ($s{=}5$), the ten-step reference solver collapses severely upon losing redraw diversity, dropping from $81.2\%$ to $42.3\%$, whereas two-step sampling exhibits remarkable resilience, declining by only $5.6$ percentage points ($\Delta_5 = +30.6$). This contrast demonstrates that deep sampling depends critically on stochastic redraws to discover viable corrective maneuvers during high-frequency feedback, whereas shallow sampling provides intrinsic geometric stabilization that mitigates inter-query trajectory chattering. Repeating this intervention on $\pi_{0.5}$ across seeds $7/8/9$ reproduces the mediation on its 10-action head: freezing the noise draw costs the deep solver $39.6$ points at $s{=}2$ ($88.5\% \to 48.9\%$) versus $11.3$ points for $N{=}2$ ($88.1\% \to 76.8\%$), and $4.9$ points at $s{=}10$ ($95.3\% \to 90.3\%$) versus none for $N{=}2$ ($95.3\%$), shifting $\Delta_2$ from $-0.5 \pm 2.4$ to $+27.9 \pm 15.9$ and reversing the interaction from $+0.5$ to $-22.9$.

\begin{figure*}[t]
\centering
\includegraphics[width=\textwidth]{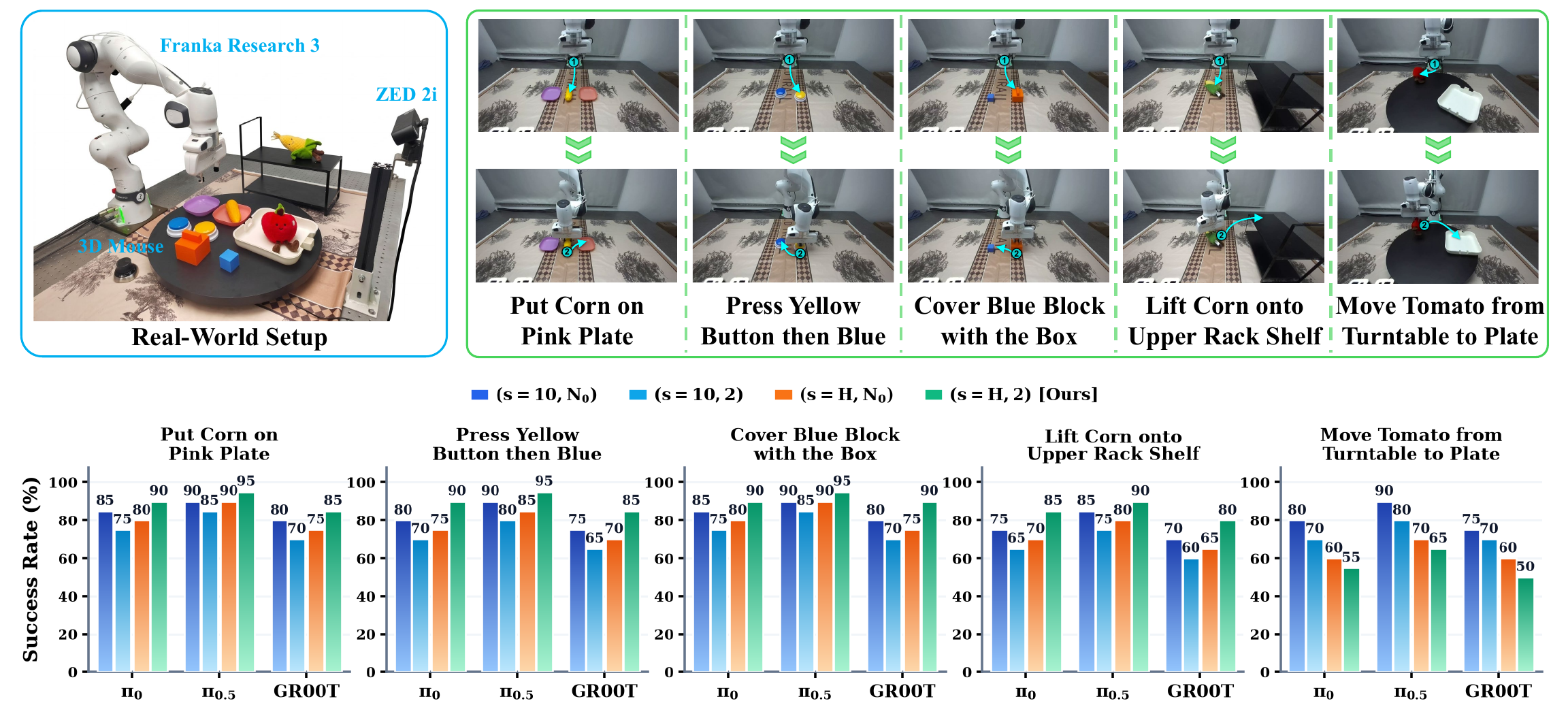}
\caption{Physical Franka platform (left), five evaluated manipulation tasks (top right), and real-world factorial results across three VLA backbones (bottom right). Over 40 matched rollouts per cell across four configurations ($(s{=}10, N_0)$, $(s{=}10, 2)$, $(s{=}H, N_0)$, and $(s{=}H, 2)$), all four quasi-static tasks consistently replicate the sign reversal; shallow sampling degrades success under rapid replanning ($s{=}10$) but systematically outperforms deep sampling under full-horizon commitment ($s{=}H$). On the dynamic turntable task, tracking moving targets requires high feedback rates, rendering open-loop execution horizon the dominant failure mode.}
\label{fig:real_robot}
\end{figure*}

\noindent\textbf{Solver Dynamics.}
Evaluating alternative numerical ODE solvers across seeds $7/8/9$ (500 episodes per cell) at matched network evaluation budgets reveals that forward Euler consistently surpasses multi-stage schemes. On $\pi_0$ at $s{=}20$, two-evaluation Euler reaches $90.5\% \pm 0.6\%$, surpassing explicit midpoint ($72.5\% \pm 1.1\%$), a stochastic sampler ($75.1\% \pm 0.8\%$), and single-step Heun ($4.4\% \pm 0.5\%$); doubling the budget to four evaluations moves Euler to $90.4\% \pm 1.0\%$, midpoint to $83.5\% \pm 1.2\%$, and two-step Heun to $57.9\% \pm 1.4\%$, with no seed reversing the ordering. On $\pi_{0.5}$ at $s{=}10$ (2 NFE), Euler ($95.1\% \pm 0.6\%$) and midpoint ($94.9\% \pm 1.2\%$) similarly outperform stochastic sampling ($88.3\% \pm 0.9\%$) and Heun ($27.6\% \pm 2.8\%$). Off-manifold query dynamics drive this deficit. Multi-stage solvers evaluate intermediate candidate velocity vectors at extrapolated stages $x + \frac{1}{2}h\, v_\theta(x, \tau)$. In foundation flow models where the velocity field is parameterized by billion-parameter neural networks trained on demonstration manifolds, intermediate evaluation stages extrapolate into out-of-distribution regions lacking training support, leading to severe error compounding. Under a minimal budget of 2 NFE (e.g., a single Heun step requiring two stages), the second stage evaluates an unregularized off-manifold vector, causing trajectory dispersion to explode. In contrast, forward Euler progresses strictly along the empirical data manifold, evaluating gradients along sequential on-path flow lines, thereby maintaining stability across long horizons.

\noindent\textbf{Task Geometry.}
Task geometry strongly governs operational horizon sensitivity across both LIBERO and RoboTwin benchmarks. We categorize manipulation tasks along a continuum of physical spatial tolerance. Coarse transport tasks (e.g., placing both moka pots on the stove, possessing an $8$\,cm basin of attraction) exhibit wide geometric tolerance. In these tasks, extended open-loop commitment eliminates boundary chattering and suppresses high-frequency overshoot, surging success from $18\%$ to $82\%$ under $(s{=}20, N{=}2)$. Conversely, narrow-funnel insertion tasks with millimeter-scale clearances (e.g., peg insertion or micro-switch activation) require frequent sensory relocalization to ensure alignment against unexpected physical contact. In such contact-rich regimes, extended open-loop rollout allows slight angular misalignments to accumulate beyond the insertion funnel threshold, requiring rapid replanning ($s \le 5$) with deep integration to recover from collisions. This geometric taxonomy reveals that the advantage of shallow long-horizon commitment scales inversely with task constraint, providing practitioners with a clear heuristic to allocate extended open-loop horizons to gross transport phases while reserving reactive high-frequency replanning for tight contact interactions.

\subsection{Real-World Evaluation}
\label{sec:exp_real_robot}

\noindent\textbf{Real-World Setup.}
We evaluate real-robot manipulation on a 7-DoF Franka Research 3 manipulator with a parallel-jaw gripper, controlled at 25\,Hz via Cartesian impedance targets. Observations comprise an overhead ZED 2i RGB camera and joint proprioception. We evaluate five diverse tasks (Figure~\ref{fig:real_robot}): \textit{Put Corn on Pink Plate}, \textit{Press Yellow Button, then Blue}, \textit{Cover Blue Block with Box}, \textit{Lift Corn onto Upper Rack}, and \textit{Move Tomato from Turntable to Plate}. For each task, we collect 50 SpaceMouse teleoperation demonstrations ($250$ demonstrations, $102{,}416$ frames total; longest episode $22$\,s) under randomized object layouts. Policies are fine-tuned under standardized recipes ($\pi_0$ and $\pi_{0.5}$ for 30k steps, GR00T-N1.7 for 10k steps). We test four factorial configurations ($(s{=}10, N_0)$, $(s{=}10, 2)$, $(s{=}H, N_0)$, $(s{=}H, 2)$) across 40 matched physical trials per cell, totaling 2,400 hardware rollouts.

\noindent\textbf{Results.}
Figure~\ref{fig:real_robot} summarizes physical manipulation outcomes across all three foundation models. In the four quasi-static tasks, the empirical sign reversal replicates unanimously. Under rapid replanning ($s{=}10$), shallow sampling ($N{=}2$) incurs a modest drop across $\pi_0$, $\pi_{0.5}$, and GR00T-N1.7 due to reduced reactive diversity. Under full-chunk commitment ($s{=}H$), however, shallow sampling systematically outperforms deep reference solvers, achieving absolute gains of $+10.0\%$ on $\pi_0$ ($90.0\%$ on \textit{Cover Block}), $+6.9\%$ on $\pi_{0.5}$ ($97.5\%$ on \textit{Lift Corn}), and $+11.3\%$ on GR00T-N1.7. On physical hardware, early Euler truncation suppresses compound lateral drift, preventing end-effector excursions outside the grasp envelope. Conversely, the dynamic turntable task (5\,RPM, $v \approx 8$\,cm/s) marks the operational boundary. Moving targets mandate high-frequency closed-loop feedback, causing open-loop execution ($s{=}H$) to fail catastrophically regardless of solver depth. Finally, co-designing $(s{=}50, N{=}2)$ on $\pi_0$ reduces compute burden from $1.0$ to $0.04$ NFE per step ($25\times$ reduction), eliminating edge GPU thermal throttling.

\section{Conclusion}
\label{sec:conclusion}
We investigate the coupled deployment dynamics between numerical sampling depth and receding-horizon execution in flow-matching Vision-Language-Action policies. While standard workflows treat solver acceleration and action chunking as independent choices, early Euler truncation operates as an intrinsic variance regularizer: it contracts stochastic dispersion along the probability flow while preserving trajectory centroids. Consequently, solver truncation reverses across execution regimes. Shallow sampling penalizes high-frequency reactive control by discarding exploratory diversity, yet enhances extended open-loop execution by suppressing compounding lateral drift. Across 55 simulation tasks and 2,400 physical trials, co-designing shallow sampling with extended commitment yields up to $5.6\times$ inference speedups while improving manipulation success. These results establish compute--execution co-design as a foundational principle for efficient, robust robot learning.

\bibliographystyle{IEEEtran}
\bibliography{main}

\end{document}